\documentclass[a4paper,twoside]{article}

\usepackage{epsfig}
\usepackage{subfigure}
\usepackage{calc}
\usepackage{amssymb}
\usepackage{amstext}
\usepackage{amsmath}
\usepackage{amsthm}
\usepackage{multicol}
\usepackage{pslatex}
\usepackage{apalike}
\usepackage{SCITEPRESS}     

\begin{document}

\title{Head Detection with Depth Images in the Wild}

\author{\authorname{Diego Ballotta, Guido Borghi, Roberto Vezzani and Rita Cucchiara}
\affiliation{Department of Engineering "Enzo Ferrari" \\ University of Modena and Reggio Emilia, Italy}
\email{\{name.surname\}@unimore.it}
}



\abstract{
Head detection and localization is a demanding task and a key element for many computer vision applications, like video surveillance, Human Computer Interaction and face analysis. The stunning amount of work done for detecting faces on RGB images, together with the availability of huge face datasets, allowed to setup very effective systems on that domain. However, due to illumination issues, infrared or depth cameras may be required in real applications.   
In this paper, we introduce a novel method for head detection on depth images that exploits the classification ability of deep learning approaches. In addition to reduce the dependency on the external illumination, depth images implicitly embed useful information to deal with the scale of the target objects.
Two public datasets have been exploited: the first one, called \textit{Pandora}, is used to train a deep binary classifier with face and non-face images. The second one, collected by \textit{Cornell University}, is used to perform a cross-dataset test during daily activities in unconstrained environments. 
Experimental results show that the proposed method overcomes the performance of state-of-art methods working on depth images.
}

\onecolumn \maketitle \normalsize \vfill


\section{\uppercase{Introduction}}
Human head detection is a traditional computer vision research field, and in last decades many efforts have been conducted to find competitive and accurate methods and solutions. This task is a fundamental step for many research fields based on faces, such as face recognition, attention analysis, pedestrian detection, human tracking, to develop real world applications in contexts such as video surveillance, autonomous driving, behavior analysis and so on.\\
Variations in appearance and pose, the presence of strong body occlusions, lighting condition changes and cluttered backgrounds made head detection a very challenging task in wild contexts. Moreover, the head could be turned away from the camera or could be captured in a far-field.\\
Most of the current research approaches are based on images taken by conventional visible-lights cameras -- \textit{i.e.} RGB or intensity cameras -- and only few works tackle the problem of head detection in other types of images, like \textit{depth images}, also known as depth maps or range images. 
Recently, the interest on the exploitation of depth images is increasing, thanks to the wide spread of low cost, ready-to-use and high quality depth acquisition devices, \textit{e.g.} \textit{Microsoft Kinect} or \textit{Intel RealSense} devices. Furthermore, these recent depth acquisition devices are based on infrared light and not lasers, so they are not dangerous for humans and can be used in human environment without particular limitations.\\
\begin{figure}[t!]
\centering
\includegraphics[width=1\linewidth]{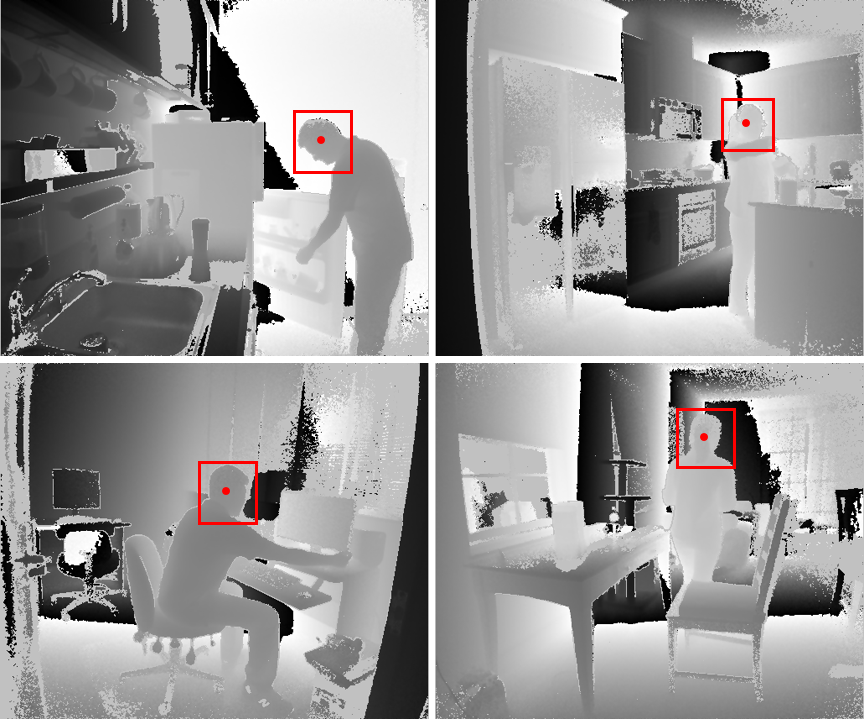}
\caption{Head detection results (head center and head bounding box depicted as a red dot and rectangle, respectively) on sample depth frames taken from the \textit{Watch-n-Patch} dataset, that contains kitchen (first row) and office (second row) daily activities acquired by a depth sensor (\textit{i.e. Microsoft Kinect One}). } 
\label{fig:headdetection} 
\end{figure}
Some real situations, dominated for instance by even dramatic lighting changes or the absence of the light source, strictly impose the exploitation of light-invariant vision-based systems. An example can be represented by a driver's behavior monitoring system, that is required working during the day and the night, with different weather conditions and road contexts (\textit{i.e.} clouds, tunnels), in which conventional RGB images could be not available or have a poor quality. Infrared or depth cameras may help to achieve light invariance.\\
%
\begin{figure*}[t!]
\centering
\includegraphics[width=1\linewidth]{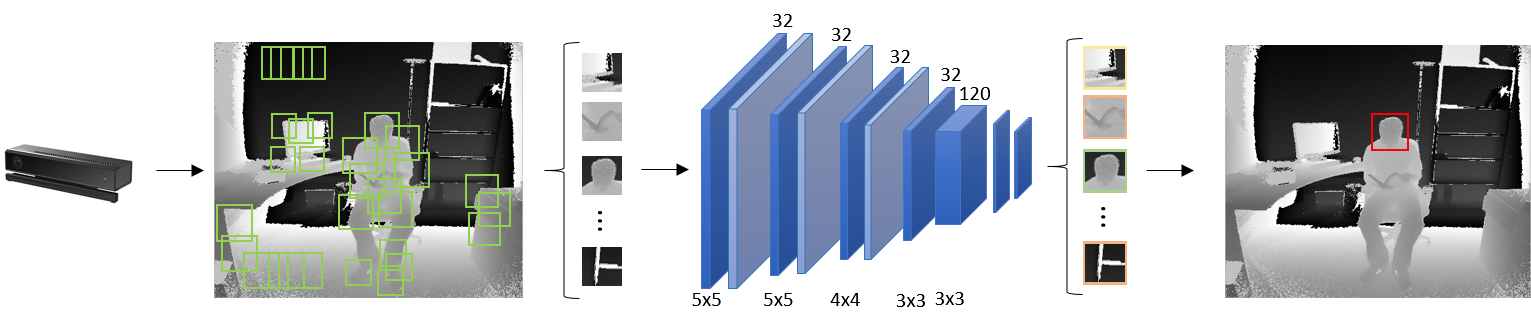}
\caption{Overall schema of the proposed framework. From the left, depth frames are acquired with a depth device (like \textit{Microsoft Kinect}), then patches are extracted and sent to the CNN, a classifier that is able to predict if a candidate patch contains a head or not. Finally, the position of the head patch is recovered, to find the coordinates into the input frame. To facilitate the visualization, 16 bit depth images are shown as 8 bit gray level images and only few extracted patches are reported.} 
\label{fig:overall} 
\end{figure*}
\noindent Besides, head detection methods based on depth images have several advantages over method based on 2D information. In particular, 2D based methods generally suffer the complexity of the background and when subject's head has not a consistent color or texture. Finally, depth maps can be exploited to deal with the scale of the target object in detection tasks, as described below in Section \ref{patches}. \\

In this paper, we present a method that is able to detect and localize a head, given a single depth image. To the best of our knowledge, this is one of the first method that exploits both depth maps and a Convolutional Neural Network for the head detection task. The proposed system is based on a deep architecture, created to have good accuracy and to be able to classify head or non-head images. Our deep classifier is trained on a recent public dataset, \textit{Pandora} introduced in \cite{borghi17cvpr}, and the whole system is tested on an another public dataset, namely \textit{Watch-n-Patch} dataset \cite{Wu_2015_CVPR}, collected by the \textit{Cornell University}, performing a cross-dataset evaluation.\\
Results confirm the effectiveness and the feasibility of the proposed method, also for real world applications.\\

The paper is organized as follows. Section \ref{sec:related}  presents an overall description of related literature works, about head detection and also pedestrian detection. In Section \ref{sec:facedetection} the presented method is detailed: in particular, the architecture of the network and the pre-processing phase for the input data are described. In Section \ref{sec:results} experimental results are reported and also a description of datasets exploited to train and test the presented method. Finally, Section \ref{sec:conclusions} draws some conclusions and includes new directions for future work.

\section{\uppercase{Related Work}} \label{sec:related}
As described above, most of head detection methods proposed in the literature are based on intensity or RGB images. This is the case of the well-known \textit{Viola-Jones} object detector \cite{viola2004robust}, where \textit{Haar} features and a cascade classifier (AdaBoost \cite{freund1995desicion}) are exploited to develop a real time and a robust face detector. A specific set of features has to be collected to handle the variety of head poses. Besides, solutions based on SVM \cite{osuna1997training} and Neural Networks \cite{rowley1998neural} have been proposed to tackle the problem of face or head detection on intensity images.\\

Very few works present approaches for head detection only based on depth images. The recent work of Chen \textit{et al.} \cite{chen2016exploring} presents a novel head descriptor to classify, through a \textit{Linear Discriminant Analysis} (LDA) classifier, pixels as head or non-head. Depth values are exploited to eliminate false positives of head centers and to cluster pixels for final head detection. In the work of Nghiem \textit{et al.} \cite{nghiem2012head}, head detection is conducted on 3D data as first step for a fall detection system. This method detects only moving objects through background subtraction and all possible head positions are searched on contour segments. Then, modified HOG features \cite{dalal2005histograms} are computed directly on depth data, to recognize people in the image. Finally, a SVM \cite{cortes1995support} is exploited to create a head shoulder classifier. Even if the presence of other recent works that exploit CNNs with depth data \cite{venturelli17visapp,frigieri17iciap,borghi17iv,borghi16icpr}, we believe that this paper proposes a novel approach for head detection on only depth maps. \\

In some works, only head localization task is addressed, that is the ability to localize the head into the image, assuming the presence of at least one head in the input image. In these cases, it is frequently supposed that the subject is frontally placed in respect to the acquisition device. This is the case of \cite{fanelli2011real}, where depth image patches are used to directly estimate head location and orientation at the same time with a regression forests algorithm. Also in \cite{borghi17cvpr} the head center is predicted through a regressive Convolutional Neural Network (CNN) trained on depth frames, supposing the user's head is present and at least partially centered in frames acquired. In both cases, authors assume also that head is always visible on the top of the moving human body. \\
Head detection shares some common aspects with face recognition and pedestrian detection tasks and this is why face detection methods often cover the case of human or pedestrian detection. Moreover, most head detector rely on the assumption to find shoulder joints to locate head into the input image.\\
Gradient-based features such as HOG \cite{dalal2005histograms}, EOH \cite{levi2004learning} are generally exploited for pedestrian detection in gray level images. Techniques to extract scale-invariant interesting points in images are also exploited (SIFT, \cite{lowe1999object}). Other local features, like \textit{edgelets} \cite{wu2005detection} and \textit{poselets} \cite{bourdev2009poselets}, are used for highly accurate and fast human detection.\\ 
Recently, due to the great success of deep learning approaches, several methods based on CNN are proposed \cite{zhu2012face} to perform face detection, pose estimation and landmark localization, but only with intensity images.\\
A deep learning based work is presented in \cite{vu2015context}, in which a context-aware method based on local, global and pairwise deep models is used to detect person heads in RGB images. In \cite{xia2011human} a human detector based on depth data and a 2D head contour model and 3D head surface model is presented. In addition, a segmentation scheme to extract the entire body and a tracking algorithm based on detection results are proposed. \\
A multiple human detection method in depth images is presented in \cite{khan2016multiple} and is based on a fast template matching algorithm; results are verified though a 3D model fitting technique. Then, human body is extracted exploiting morphological operators to perform a segmentation scheme. In \cite{ikemura2011real} a method for detecting humans by relational depth similarity features based on depth information is presented: integral depth images and \textit{AdaBoost} classifier are exploited to achieve good accuracy and real time performance. \\
Shotton \textit{et al.} proposes in \cite{shotton2013real} a method based on randomized decision trees to quickly and accurately predict the 3D positions of body joints from a single depth image (also the head is included) and no temporal information are exploited.

\section{\uppercase{Face Detection}} \label{sec:facedetection}
A depiction of the overall framework is shown in Figure \ref{fig:overall}. Given a single depth image as input, a set of square patches, the head candidates, is extracted and fed to a CNN based classifier, which predicts if the patch contains a human head or not.\\
Positively classified patches are further analyzed to refine the classification and reject non-maxima proposals. The final output of the system are the $\{x_i,y_i\}$ coordinates of the detected face centers and the corresponding bounding box sizes $\{w_i,h_i\}$. 

\subsection{Depth camera and data pre-processing}
Some features of the presented approach are related to the acquisition device used to collect depth maps. 
Thus, before describing the method, let us provide a brief introduction to the acquisition devices that are usually used to collect depth frames.\\ 
Both \textit{Pandora} and \textit{Watch-n-Patch} datasets exploit the second generation \textit{Microsof Kinect One} device, a \textit{Time-of-Flight} (ToF) depth camera. Thanks to its infrared light, it is able to measure the distance to an object inside the scene, by measuring the time interval taken for infrared light to be reflected by the object in the scene.
\textit{Kinect One} is able to acquire data in real time (30 fps), with a range starting from 0.5 to 8 meters, but best depth information are available only up to 5 meters. All distance data are reported in millimeters. 
The sensor provides depth information as a two dimensional array of pixels (depth maps), like a gray-level image. Since each pixel represents the distance in millimeters from the camera, depth images are represented in 16 bit. For this reason, in our system we use 16 bit input images, and depth values are converted in standard 8 bit values (0 to 255) only for visual inspection or representation.\\
Due to the nature of ToF sensors, noise is frequently present in acquired depth images The noise is visible as dots with zero value (random black spots) and therefore input depth images are pre-processed to remove these values through a $3 \times 3$ median filter. \textit{Kinect One} is able to simultaneously acquire RGB and depth images with a spatial resolution of $1920 \times 1080$ and $512 \times 424$, respectively. In our approach, we work only on depth frames with full resolution.

\subsection{Patch extraction} \label{patches}
Apart from the pre-processing stage, the extraction of candidate patches is the first step of the system. Without any additional information or constraint, a face can be located everywhere in the image with an unknown scale. As a result, a complete set of face candidates can be obtained with a \textit{sliding-window} approach performed at different scales. Empirical rules can be used to reduce the cardinality of the candidate set, for example by adopting pyramidal procedures or by reducing the number of tested scales or the overlapping between consecutive spatial samples. As a drawback, the precision of the method will be degraded. \\

\begin{figure}[t!]
\centering
\subfigure[]{\includegraphics[width=1\columnwidth]{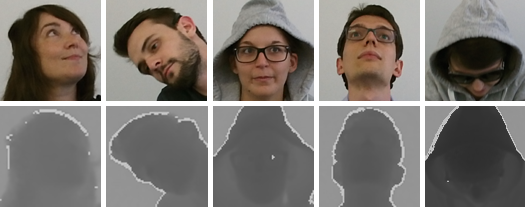}} 
\subfigure[]{\includegraphics[width=1\columnwidth]{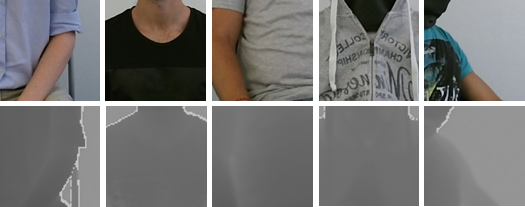}}
\caption{(a) Examples of extracted head patches for the training phase of our classifier. As shown, turned head, prescription glasses, hoods and other type of garments and occlusions can be present. (b) Examples of extracted non-head patches, including body parts and other details. All reported candidates are taken from \textit{Pandora} dataset. For a better visualization, depth images are contrast-stretched and resized to the same size.} 
\label{fig:candidates} 
\end{figure}

Differently from appearance images, depth maps embed the distance of the object to the cameras. Calibration parameters can be exploited to estimate the size of a head in the image given its distance from the camera and vice versa. Candidate patches can be early rejected if the above mentioned constraint is not satisfied. \\
More precisely, for each candidate head center $p=\{x,y\}$ within the image, the distance $D_p$ of the object in that position is recovered by averaging the depth values over a small square neighborhood of radius $K$ around $\{x,y\}$. 
Given the average size of a real human head and the calibration parameters, the corresponding image size is computed and used as the unique tested scale for that center position $\{x,y\}$. Analytically, the width and the height of the extracted candidate patch ($w_p, h_p$) centered in $p=\{x,y\}$ are computes as follow: 
\begin{equation}
w_p = \frac{f_x \cdot R}{D_p} \quad h_p = \frac{f_y \cdot R}{D_p}
\label{eq:crop}
\end{equation}
where $f_x, f_y$ are the horizontal and the vertical focal lengths of the acquisition device (expressed in pixels) and $R$ is a constant value representing the average width of a face ($200$ mm in our experiments). To further reduce the amount of candidates, the head center positions are sampled each $K$ pixels. Thus, given a input image of width and height $(w_i, h_i)$, the total number of extracted patches is computed as follow:
\begin{equation}
\#_{patches} = \frac{w_i \cdot h_i}{K^2}.
\label{eq:number}
\end{equation}
Patches smaller than $15 \times 15$ pixels are discarded, since they correspond to background objects.\\
With this procedure we obtain two major benefits: we do not need to implement any multiple-scale approach, like in \cite{viola2004robust}, so the processing overhead is reduced; the second benefit is that we are able to extract square candidates that well fit a person head, if present, and  only a minor part of the background, as depicted in Figure \ref{fig:candidates} (a).\\ 
All the patches are then resized to $64 \times 64$ pixels.\\
Supposing the head in a central position in the patch, even if we include minor parts of the background, we maintain only foreground, the person head, setting to 0 all the depth values into the patch greater than $D + l$, where $l$ is the general amount of space for a head and $D$ is the same value computed above. Final patch values are then normalized to obtain values between $[-1, 1]$. This normalization is also required by the specific activation function that is adopted in the network architecture (see Section \ref{sec:network}) and it is a fundamental step to improve CNN performance, as described in \cite{krizhevsky2012imagenet}. 

\begin{table*}[t!]
\centering
\caption{Comparison between different baselines of our method. In particular, the pixel area $K$ is varied, influencing the performance in terms of detection rate and frames per second. True positives, \textit{Intersection over Union} (IoU) and \textit{frames per second} (fps) are reported. We take the best result for the comparison with the state-of-art.}
\label{tab:kernelcomparison}
\begin{tabular}{c|ccc}
\hline
\textbf{Pixel Area (K)} & \textbf{True Positives} &\textbf{Intersection over Union} &\textbf{Frames per Second}  \\ \hline \hline
$3 \times 3$		&\textbf{0.883}	&\textbf{0.635} &0.103\\
$5 \times 5$		&0.873	&0.631 &0.235\\
$9 \times 9$		&0.850	&0.618 &0.655\\
$13 \times 13$		&0.837	&0.601 &1.273\\
$17 \times 17$		&0.786	&0.569 &2.124\\
$21 \times 21$		&0.748	&0.528 &2.622\\
$45 \times 45$		&0.376	&0.009 &\textbf{12.986}\\

\hline
\end{tabular}
\end{table*}

\subsection{Network Architecture} \label{sec:network}
Taking inspiration from \cite{krizhevsky2012imagenet}, we adopt a shallow network architecture to deal with computation time and system performance. Another relevant element is the lack of publicly available annotated depth data, that forced us to adopt deep models with a limited number of internal parameters as depicted in Figure \ref{fig:overall}. The proposed network takes as input depth images of $64 \times 64$ pixels. There are 5 convolutional layers, where the first four have 32 filter with size of $5 \times 5$, $4 \times 4$ and $3 \times 3$ respectively, and the last one has 128 filters, with size of $3 \times 3$. Max-pooling is conducted only on the first three convolutional layers, due to the limited size of input images. Three fully connected layers are then added with 128, 84 and 2 neurons respectively. We adopt the hyperbolic tangent (\textit{tanh}) as activation function in all layers:
\begin{equation}
tanh(x) = \frac{2}{1 + e^{-2x}}-1
\end{equation}
in this way network is able to map input $[-\infty, +\infty] \rightarrow [-1, +1]$. In the last fully connected layer we adopt the \textit{softmax} activation to perform the classification task.\\
We exploit \textit{Adam} solver \cite{kingma2014adam}, with an initial learning rate set to $10^{-4}$, to resolve back-propagation and automatically adjust the learning rate values during the training phase. We exploit data augmentation technique to avoid over-fitting phenomena \cite{krizhevsky2012imagenet} and increase the number of training data: each input image is flipped, so the final number of input images is doubled. The categorical cross-entropy function as been used as loss.

\section{\uppercase{Results}} \label{sec:results}
In this section, experimental results of the proposed method are presented. In order to evaluate its performance, we use two public and recent datasets, \textit{Pandora} for the patch extraction and the network training part and the second dataset, \textit{Watch-n-Patch} for the testing phase, the same used in \cite{chen2016exploring}. Experimental results for \cite{nghiem2012head} on \textit{Watch-n-Patch} dataset are taken from \cite{chen2016exploring}.\\

Generally, head detection task with depth images task lacks of the availability of publicly datasets, specifically created for face or head detection in wild contexts. Several datasets containing both depth data and visible human heads were collected in this decade, \textit{e.g.} \cite{fanelli2011real,baltruvsaitis20123d,bagdanov2011florence}, but they present some issues, for example they are not deep learning oriented, due to their very limited number of annotated samples. Moreover, subjects are often still, perform too static actions, and frontal face the acquisition device. Besides, we consider only dataset with \textit{Time-of-Flight} data, that contains frames with higher quality and depth measures accuracy as described in \cite{sarbolandi2015kinect}, in respect with structured-light sensors (like the first version of \textit{Kinect}).\\
As mentioned above, we exploit \textit{Pandora} dataset to generate patches of head and non-head, based on skeleton annotations, to train our CNN. Non-head candidates are extracted randomly sampling depth frames, excluding head areas. An example of extracted head patches and non-head patches used for the training phase is reported in Figure \ref{fig:candidates}. Due to \textit{Pandora} dataset features, heads with extreme poses, occlusions and garments can be present.  

\begin{table*}[t!]
\centering
\caption{Results on \textit{Watch-n-Patch} dataset for head detection task, reported as True Positives and False Positives. The proposed method largely overcomes literature competitors.}
\label{tab:comparison}
\begin{tabular}{c|cc|cc}
\hline
\textbf{Methods} &\textbf{Classifier} &\textbf{Features} & \textbf{True Positives} &\textbf{False Positives}   \\ \hline \hline
\cite{nghiem2012head}		&SVM		&modified HOG	&0.519	&0.076 \\
\cite{chen2016exploring}	&LDA 		&depth-based head descriptor	&0.709	&0.108 \\
\textbf{Our}				&\textbf{CNN} &\textbf{deep}		&\textbf{0.883}	&\textbf{0.077} \\
\hline
\end{tabular}
\end{table*}

\subsection{Pandora dataset}
\textit{Pandora} dataset is introduced in \cite{borghi17cvpr}. It is acquired with \textit{Microsoft Kinect One} device and is specifically created for the head pose estimation task. It is deep oriented, since it contains about 250k frames divided in 110 sequences of 22 subjects (10 males and 12 females). It is a challenging dataset, subjects can vary their head appearance wearing prescription glasses, sun glasses, scarves, caps and manipulate smart-phones, tablets and plastic bottles that can generate head and body occlusions. It is a useful dataset to extract patches due to the presence of head with various poses and appearance. Skeleton annotations facilitate the extraction of head and non-head patches.

\begin{figure}[b!]
\centering
\includegraphics[width=1\columnwidth]{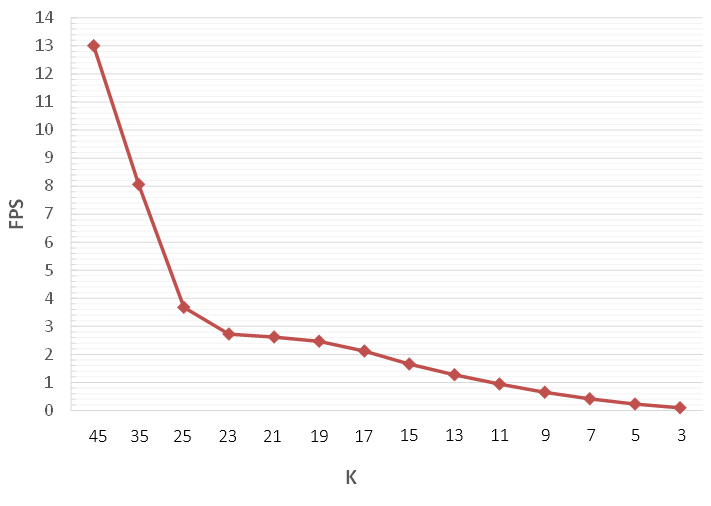}
\caption{Correlation between the pixel area $K$ and speed performance, in terms of \textit{fps}, of the proposed method. As expected, detection rate is low with high speed performance and vice versa.} 
\label{fig:grafico} 
\end{figure}

\subsection{Watch-n-Patch dataset}
Wu \textit{et al.} introduces this dataset in \cite{Wu_2015_CVPR}. It is created with the focus on modeling human activities, comprising multiple actions in a completely unsupervised setting. Like \textit{Pandora}, it is collected with \textit{Microsoft Kinect One} sensor for a total length of about 230 minutes, divided in 458 videos. 7 subjects perform human daily activities in 8 offices and 5 kitchens with complex backgrounds, in this way different views and head poses are guaranteed. \\
Moreover, skeleton data are provided as ground truth annotations. Even if this dataset is not explicitly created for head detection task, it is a useful dataset to test head detection system on depth images, thanks to its variety in poses, actions, subjects and background complexity.

\begin{figure*}[t!]
\centering
\includegraphics[width=0.99\linewidth]{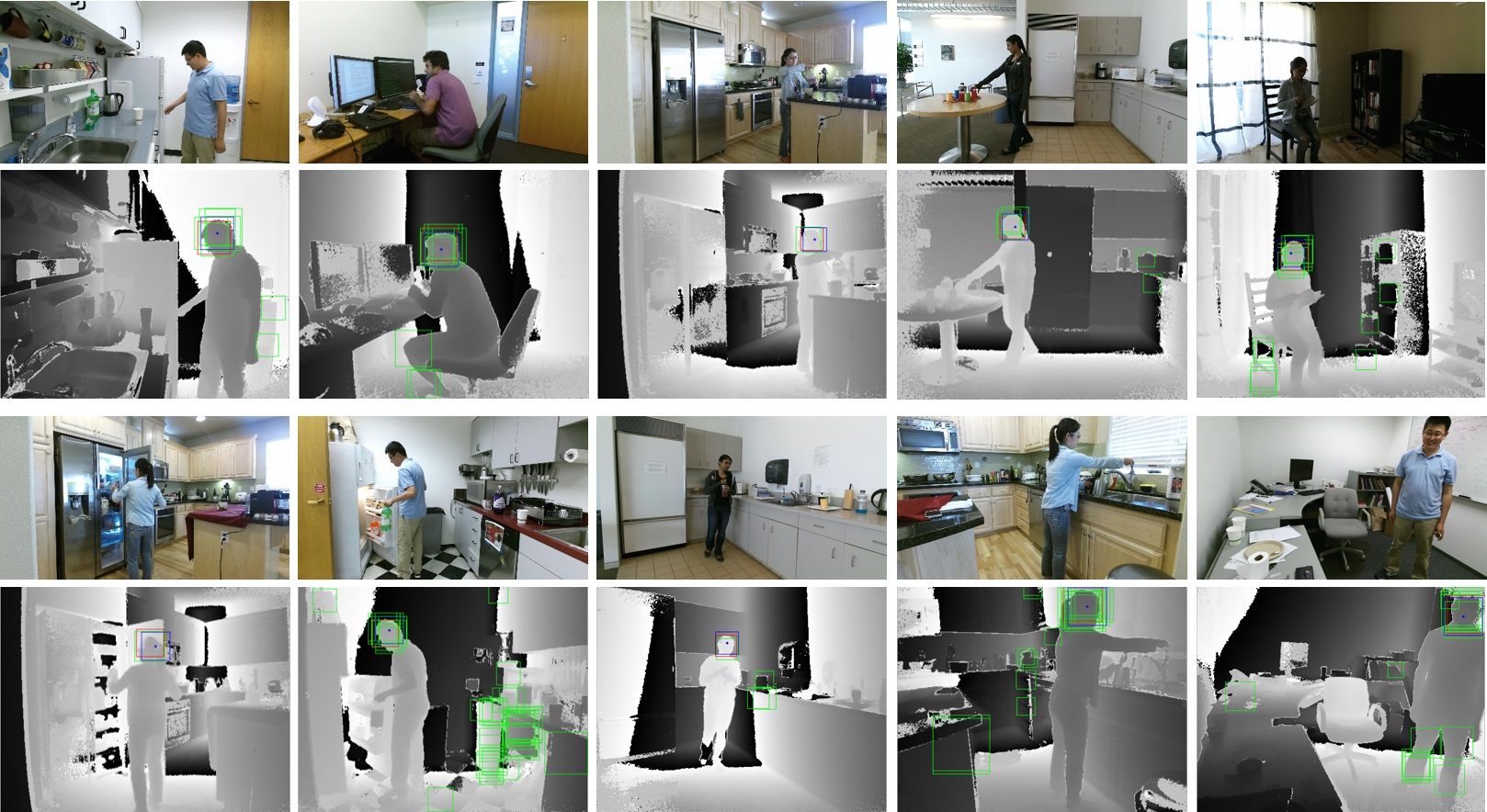}
\caption{Example outputs of the proposed system. RGB frames are reported in the first and third rows, while in the second and the last rows are depicted the correspondent depth maps. Our prediction is reported as blue rectangle, also ground truth (red rectangle) and some other patch candidates (green) are shown. Sample frames are collected from \textit{Watch-n-Patch} dataset. [best in colors]}
\label{fig:output} 
\end{figure*} 

\subsection{Experimental results}
First, we investigate performance of the proposed system varying the size of $K$, the pixel area used to compute the average distance $D$ (see. Section \ref{patches}) between a point in the scene and the acquisition camera. The size of the pixel area $K$ affects both the computation time, due to the final number of patches generated on input images (see Equation \ref{eq:number}), and the head detection rate: a bad or corrupted estimation of $D$ generate low quality patches and could compromise CNN classification performance.

\noindent In our experiments, for a fair comparison with \cite{chen2016exploring}, we report our results as true positives number of head detected. We consider also \textit{Intersection over Union} (IoU) metric and \textit{frames per second} (fps) value to check time performance of the proposed system. A head is correctly detected and localized (True Positive) only if
\begin{equation}
IoU(A, B) > \tau 
\end{equation}
\begin{equation}
IoU(A, B) = \frac{|A \cap B |}{|A \cup B| - |A \cap B|}
\end{equation}
where $A, B$ are ground truth and predicted head bounding boxes, respectively. According to \cite{chen2016exploring}, $\tau = 0.5$.

\noindent If a patch is incorrectly detected as a head by CNN, it creates one false positive and one false negative. As above reported, we include also computation time, that includes the part of patch extraction and the part of CNN classification. Results are reported in Table \ref{tab:kernelcomparison} . As expected, system accuracy decreases and time computation increases with smaller $K$ size. Thus, the size of the pixel area $K$ can be set based on the type of final application in which head detection is necessary, where can be preferred accuracy or speed performance. Tests have been carried on a \textit{Intel i7-4790} CPU (3.60 GHz) and with a \textit{NVIDIA Quadro k2200}. The deep model has be implemented and tested with \textit{Keras} \cite{chollet2015keras} with \textit{Theano} \cite{2016arXiv160502688short} back-end.\\
Since we proposed a head detection only based on depth images, we compare our method with two state-of-art head detection systems based on depth data: the first one has been introduced by Chen \textit{et al.} in \cite{chen2016exploring}; the second one is a system for fall detection proposed in \cite{nghiem2012head}. Comparisons with methods based on intensity images or hybrid approaches are out of the scope of this paper. 
Evaluations with other depth-based head localization methods present in the literature reported in Section \ref{sec:related} \cite{fanelli2011real,borghi17cvpr} are not feasible, since a specific context for acquired scenes is strictly required, \textit{i.e.} a person facing the acquisition device, with only the upper body part visible.  \\
For the experimental comparison, we exploit as ground truth the skeleton data provided with the \textit{Watch-n-Patch} dataset. 
Results are reported as the total number of True Positive and False Positive of head detection, given a subsequence of the \textit{Watch-n-Patch} dataset (2785 images). For the sake of fairness,  we exploited the same subsequences used in \cite{chen2016exploring,nghiem2012head}, as stated by the authors. This subset has been chosen by authors due to the presence of scene with good background quality, required by \cite{nghiem2012head}, and the presence of people, to meet the assumption present in \cite{chen2016exploring} of existing heads in all input images. It is important to note that our approach do not rely on these two strong assumptions. All data with wrong ground truth annotations or missing depth data are discarded.\\
Table \ref{tab:comparison} reports the comparison with the state-of-art. Our method achieves better performance in terms of true positive number. In particular scenes, we achieve a 100\% of correct head detections and the cross-dataset evaluation guarantees the generalization capability of the proposed architecture. Finally, we note that in \cite{chen2016exploring,nghiem2012head} is not clearly reported how the number of false positive is computed.

\section{\uppercase{Conclusions}} \label{sec:conclusions}
A novel method to detect and localize a head from a single depth image is presented. The system is based on a Convolutional Neural Network designed to classify, like a binary classifier, candidates as head or non-head. Results confirm the feasibility and the accuracy of our method, that can be a key element for frameworks created for face recognition or behavior analysis, in environments in which light invariance is strictly required.\\
The flexibility of our approach allows the possibility of future work, that may involve the investigation of multiple head detection task in depth frames. Generally, the acquisition of new data is needed, due to the lack of specific dataset created for single and multiple head detection with depth maps. Moreover, future extensions are also related with the reduction of the computation time.

\bibliographystyle{apalike}
{\small
\bibliography{Visapp2018_HeadDetection_Ballotta}}

\end{document}